\documentclass[11pt]{article}

\usepackage[in]{fullpage}
\usepackage{bm}

\usepackage{hyperref}

\usepackage{times}
\usepackage{helvet}
\usepackage{courier}
\usepackage{graphicx}
\usepackage[cmex10]{amsmath}
\usepackage[mathscr]{euscript}
\usepackage{bm}
\usepackage{amsfonts}
\usepackage{tikz}
\usepackage[mathscr]{euscript}
\usepackage{url}

\begin{document}

\title{Semi-Markov Switching Vector Autoregressive Model-based Anomaly Detection in Aviation Systems}

\author{Igor Melnyk${}^1$ \\
melnyk@cs.umn.edu
\and
Arindam Banerjee${}^1$\\
banerjee@cs.umn.edu
\and
Bryan Matthews${}^2$\\
bryan.l.matthews@nasa.gov
\and
Nikunj Oza${}^2$\\
nikunj.c.oza@nasa.gov\\\\
${}^1$Department of Computer Science \& Engineering, Twin Cities, MN\\
${}^2$NASA Ames Research Center, Moffett Field, CA
}



\maketitle

\begin{abstract}
In this work we consider the problem of anomaly detection in heterogeneous, multivariate, variable-length time series datasets. Our focus is on the aviation safety domain, where data objects are flights and time series are sensor readings and pilot switches. In this context the goal is to detect anomalous flight segments, due to mechanical, environmental, or human factors in order
to identifying operationally significant events and provide insights into the flight operations and highlight otherwise unavailable potential safety risks and precursors to accidents.
For this purpose, we propose a framework which represents each flight using a semi-Markov switching vector autoregressive (SMS-VAR) model. Detection of anomalies is then based on measuring dissimilarities between the model's prediction and data observation. The framework is scalable, due to the inherent parallel nature of most computations, and can be used to perform online anomaly detection. Extensive experimental results on simulated and real datasets illustrate that the framework can detect various types of anomalies along with the key parameters involved.
\end{abstract}

\section{Introduction}
\label{sec:Intro}

It is estimated that by $2040$ the United States alone can expect an increase of more than $60\%$  in the commercial air traffic \cite{faa}. The anticipated air traffic growth can lead to increased congestion on the ground and air, creating conditions for possible accidents. Noting this problem, air transportation authorities are engaged in research and development of the Next Generation Air Transportation System \cite{nextgen1}, the initiative to improve the air traffic control system by increasing its capacity and utilization. A part of this effort is devoted to the processing and analysis of the air traffic flight information, also known as Flight Operations Quality Assurance (FOQA) data \cite{foqa}, to detect issues in aircraft operation, study pilot-automation interaction problems, propose corrective actions and design new training procedures.

Therefore, the objective of our work is to develop a general framework to detect operationally significant events in the flight data due to mechanical, environmental or human factors. We are specifically interested in the scenarios when the data is \emph{unlabeled}, i.e., there is no information which flights are normal and anomalous. Moreover, our interest is focused on the algorithms which can handle \emph{variable-length}, \emph{multivariate} and \emph{heterogeneous} datasets, i.e., containing multiple features which can be of mixed data type and of possibly different length across flights.

The currently deployed automated methods for the analy\-sis of FOQA data are exceedance-based approaches \cite{stma03}, \cite{nesc07}, monitoring the normal operation of the flight using predefined ranges on the parameters. Although simple and fast, this approach is limited since it examines each feature independently, ignoring potential correlations among the parameters. Moreover, since the thresholds need to be defined upfront, the method can fail to discover previously unknown abnormal events. The proposed algorithms in current research literature (see Section \ref{sec:Related}) are also not adequate since they either cannot work with the data of mixed type or their approaches for handling such data still failing to uncover the underlying problem structure and thus achieve high anomaly detection performance.

\begin{figure}[t]
\centering
\includegraphics[width=0.8\textwidth]{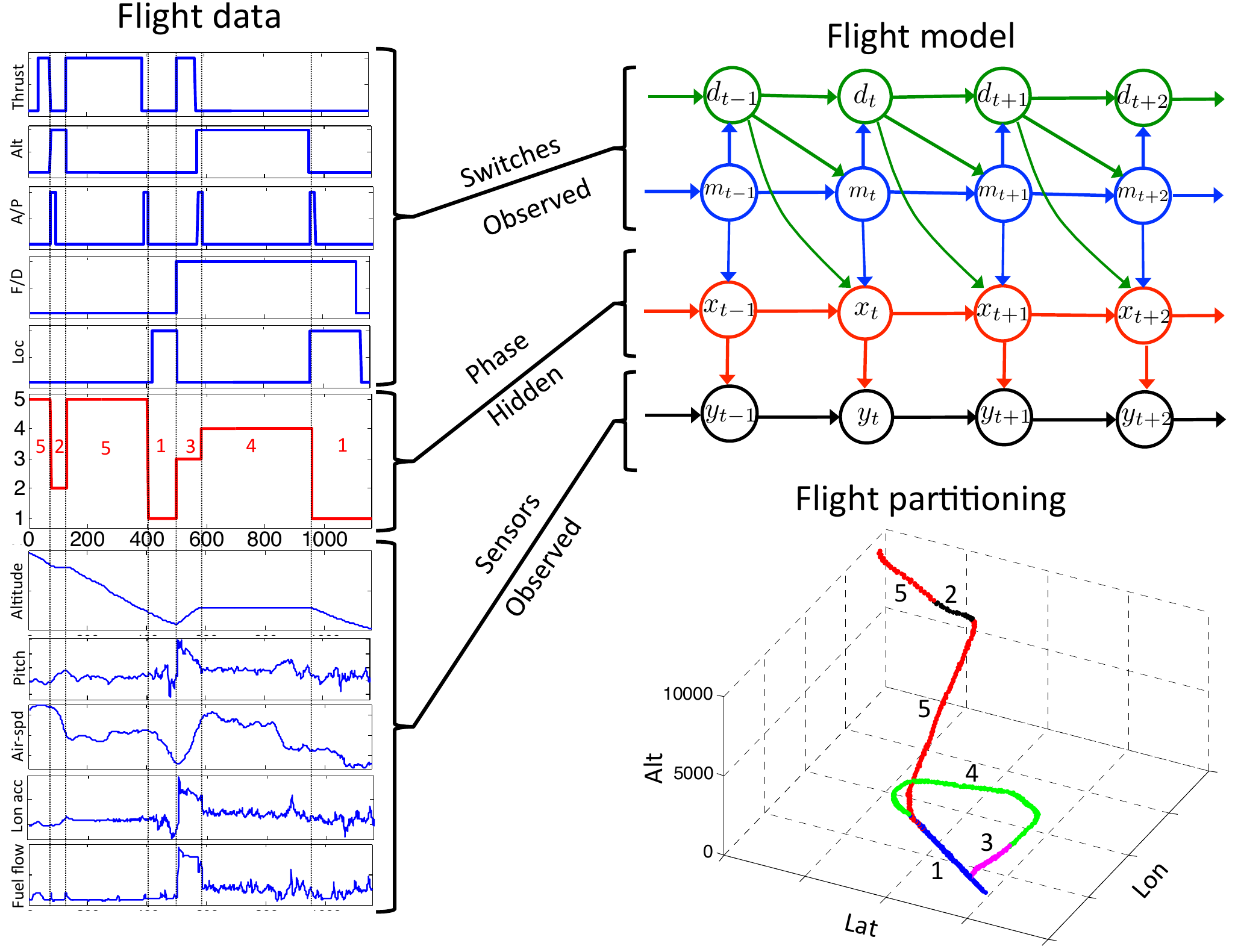}
\caption{Representing heterogeneous flight data using SMS-VAR model. Left plot shows time evolution of several pilot switches, phase and sensor measurements during aircraft's landing. Top right plot shows the proposed model to represent such data. The bottom right graph shows the trajectory of aircraft, where its path is partitioned into phases. The value of the phases is obtained after constructing SMS-VAR on such data and running Viterbi algorithm to recover most probable phase path.}
\label{fig:model}
\vspace{-5pt}
\end{figure}

To motivate our proposed approach, consider Figure \ref{fig:model}, showing a real flight data and a model to represent it. On the left we plot a part of flight related to landing and show time series of several \emph{pilot switches} (from top: thrust, altitude, autopilot, flight director, localizer) as well as some of the \emph{sensors} (altitude, pitch angle, airspeed, longitudinal acceleration, fuel flow). The switches act as controls, determining the behavior of aircraft and sensors measure the effects of the controls on the system. A combination of the switches set by a pilot determines the aircraft's behavior for a certain period of time after which a different combination of switches defines another flight period and so on. Within each flight period, called a \emph{phase} (shown in red on the left in Figure \ref{fig:model}), the aircraft's dynamics usually remains consistent and steady, while across phases the dynamics change.

As an example, consider a lower right plot in Figure \ref{fig:model}, showing aircraft's path as it descends from 10000 feet to a runway. The descent is interrupted by some event, causing it to fly back to a certain altitude, make a circle and repeat the landing. The path is partitioned into phases shown with different colors and numbers. For instance, phase 5 corresponds to a steady descent, where aircraft constantly loses altitude while maintaining its airspeed. This phases is interrupted by phase 2 of duration about 50 seconds, caused by a switch off in auto thrust system and activating hold altitude switch. In phase 2 the aircraft levels off by steadily increasing its pitch angle and losing airspeed.

Thus, if a flight is partitioned into multiple phases, determined by the pilot controls (switches), then the continuous dynamics of each phase can be represented separately by its own model. We propose to model such data with a dynamic Bayesian network - \emph{semi-Markov switching vector autoregressive} (SMS-VAR) model, shown on the right plot of Figure \ref{fig:model}. We note that our motivation comes from a rich literature of systems identification \cite{ljun98}, where a standard approach for modeling continuous system dynamics (in our case the flight's sensor measurements) is a vector autoregressive model (VAR) \cite{lutk07}. However, as we discussed above, using a single VAR model for the entire flight is inappropriate, thus we employ multiple VARs. A change from one VAR process to another, i.e., the switching behavior, is modeled with a hidden variable $x_t$, representing a flight phase.

To model the dynamics of flight switches, we convert their representation from binary categorical into discrete, i.e., at each time stamp $t$ a vector of zeros and ones (values of switches at $t$) is converted into an integer. We call the resulting variable a flight mode $m_t$ and represent it using semi-Markov model (SMM) \cite{jali13}. SMM is an extension of a simple Markov chain, allowing to model arbitrary state durations. A simple Markov chain has implicit geometric state duration distribution \cite{ross14}, causing fast state transitions and is inappropriate for our case. SMM fixes this by introducing a variable $d_t$ which controls the duration of mode $m_t$ (see Section \ref{sec:modelSpec} for more details).

To summarize, in this work we (i) propose to model multivariate, heterogeneous, variable-length flight data using SMS-VAR model (see Section \ref{sec:Model} for detailed model description and parameter learning); (ii) present anomaly detection framework which is based on the dissimilarity of one-step ahead predicted and filtered phase distributions, showing advantage over the standard likelihood-based approach; (iii) provide extensive evaluations on synthetic and real data, including 20000 unlabeled flights, showing SMS-VAR outperforming many base line algorithms and accurately detecting different types of anomalies.

The rest of the paper is organized as follows. In Section \ref{sec:Related} we review the related work on anomaly detection in aviation systems. In Section \ref{sec:Model} we formally define the SMS-VAR model and discuss the approach for its parameter estimation. In Section \ref{sec:AnomDetect} the anomaly detection methodology is presented. The experimental results are shown in Section \ref{sec:Experiments} and finally in Section \ref{sec:Conc} we conclude the paper.

\section{Related Work}
\label{sec:Related}
The problem of anomaly detection in aviation is an active area of research in the data mining community and has attracted attention of many researchers. For example, \cite{sriv12} considered a problem of detecting abnormal fuel consumption in jet engines. The method is based on using regression models to estimate consumed fuel and compare it to the observed to detect the abnormal behavior. It is a supervised approach since training requires anomaly-free data, limiting its practical application in cases when the labeled data is unavailable, as is the case in the present work.


In \cite{gomm12} the authors proposed an approach based on a specially designed linear regression model to describe the aerodynamic forces acting on an aircraft. The constructed model accounts for the flight-to-flight and aircraft-to-aircraft variability, which enables the fitting of a single model to the entire dataset. However, the postulated aerodynamics regression model requires significant domain knowledge and careful design, limiting its generalization and usage in other anomaly detection problems.

Budalakoti et.~al. \cite{bbso09} proposed a sequenceMiner, the unsupervised clustering algorithm based on longest common subsequence distance measure to detect anomalies in symbol sequences, arising from recordings of switch data. Although the algorithm can discover operationally significant safety events in switch sensors, its operation is restricted to discrete data, ignoring additional information present in the continuous flight sensors.


The work of \cite{lehl14} presented a model-based framework to identify flight human-automation issues using switches data and sensor measurements. The anomaly is identified if there is a difference between inferred intents of the automation and the observed pilot actions. The limitation of this approach is that it assumes the data is noise-free and does not account for parameter uncertainties. Moreover, it is not clear how the algorithm performs on large flight data, since the presented evaluations are limited to a few examples.

Das et.~al. \cite{dmso10} introduced MKAD, a multiple kernel learning approach for heterogeneous anomaly detection problems. The method constructs a kernel matrix as a convex combination of a kernel over discrete sequences and continuous time series. One-class SVM \cite{spss01} is then used to detect anomalies. This method was applied to the FOQA dataset \cite{mdbd13} to detect operationally significant events.  Although it showed good results, the algorithm lacks scalability since the kernel matrix has to be updated for each new test flight.

The work of Li et.~al. \cite{lshp15} proposed to detect anomalies in the flight data based on continuous and discrete features using a clustering approach, called ClusterAD. Their idea is to represent each flight as a vector, by concatenating all feature across time. After dimensionality reduction, the data is clustered based on Euclidean distance measure to identify outliers and groups of similar flights. A potential issue with this approach is a misalignment between time series from different flights. Since the size of the vectors needs to be the same for all flights, forcing such equality can introduce spurious dissimilarities, increasing false positives in the detected anomalies. Moreover, the study in \cite{dlsh12} revealed that ClusterAD does not perform well on discrete anomalies as compared to MKAD. Partially, this is due to the use of Euclidean similarity measure on heterogeneous data vectors, making it less effective in finding discrete anomalies.

Summarizing the above literature and comparing to the proposed SMS-VAR model, we can make several remarks. First, our method is unsupervised, thus it does not require a training set of labeled normal flights. Moreover, our framework works with multivariate and heterogeneous data, where each data sample can be of variable length.  As compared to data-driven methods, our framework is model-based and therefore can be computationally more efficient in the detection stage by not requiring to recompute the model for each test flight. At the same time, the model construction requires only basic knowledge about the considered parameters and can easily be extended to other anomaly detection domains.

\section{Semi-Markov Switching VAR}
\label{sec:Model}
\begin{figure*}[t]
\centering
\includegraphics[width=0.99\textwidth]{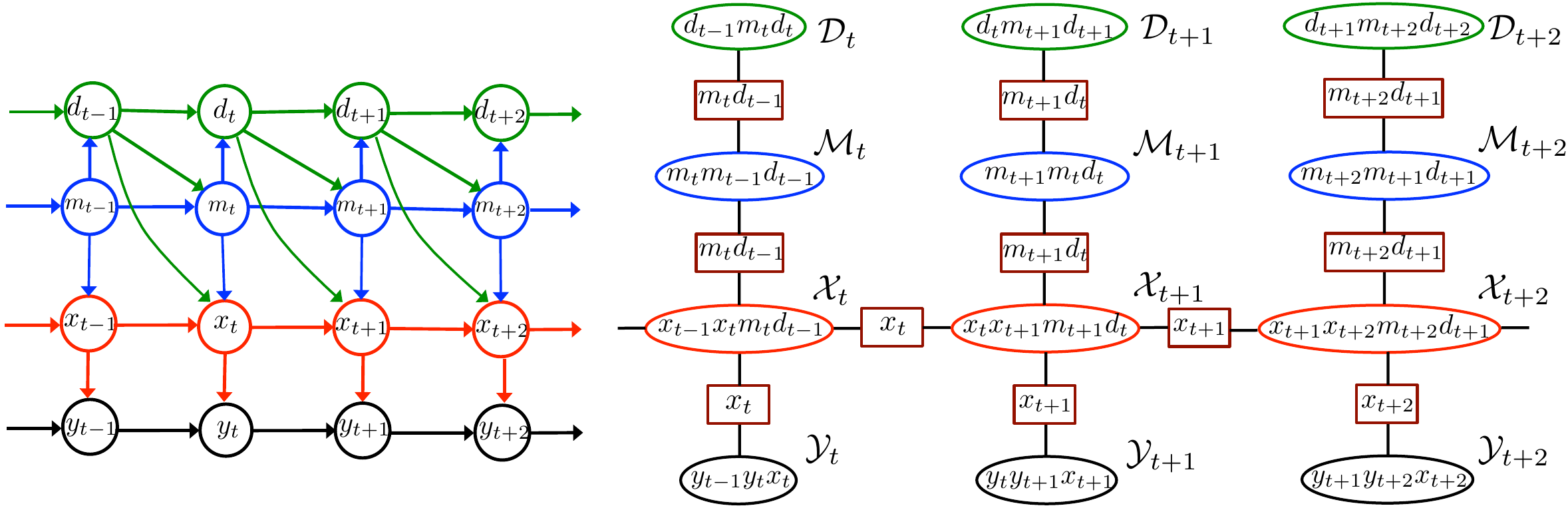}
\caption{Left: Dynamic Bayesian Network of Semi-Markov Switching Vector Autoregressive Model (SMS-VAR). Right: Junction Tree for SMS-VAR. Ovals represent graph cliques, denoted by calligraphic letters $\mathcal{D}_t, \mathcal{M}_t, \mathcal{X}_t$, and $\mathcal{Y}_t$, rectangles denote separators. Symbols within shapes show variables on which the corresponding objects depend.}
\label{fig:smsvar}
\vspace{-7pt}
\end{figure*}

In this Section we formally introduce the SMS-VAR model and show details about the parameter learning algorithm.
\subsection{Model Specification}
\label{sec:modelSpec}

In this work we propose to model the flight of an aircraft using semi-Markov switching vector autoregressive model (SMS-VAR), whose Dynamic Bayesian Network (DBN) is shown in Figure \ref{fig:smsvar}. From the graphical model perspective, SMS-VAR has four classes of variables: continuous sensor measurements $y_t \in \mathbb{R}^{n_y}$, discrete phase $x_t \in \{1,\ldots,n_x\}$, discrete mode, $m_t \in \{1,\ldots, n_m\}$ and a positive real variable $d_t \in \mathbb{Z}_{\geq 0}$, determining the time duration of the mode and phase in a particular state. The SMS-VAR model is then fully defined by specifying the probability distributions which govern the time evolution of the above four variables. In particular, the probability distribution of mode transition is modeled as
\begin{align}
p(m_t|m_{t-1},d_{t-1}) &=\begin{cases} p(m_t|m_{t-1}) & \mbox{if } d_{t-1}=1 \\
\delta(m_t, m_{t-1}) &\mbox{if } d_{t-1} > 1 \end{cases},
\label{eq:modeDist}
\end{align}
where $\delta(a,b)$ denotes the Dirac delta function: $\delta(a,b) = 1$ if $a = b$ and 0 otherwise. In essence, in this expression $d_{t-1}$ works as a down counter for mode persistence, i.e., when $d_{t-1} > 1$, the mode is forced to remain unchanged: $m_t = m_{t-1}$. On the other hand, when $d_{t-1} = 1$, a new mode state $m_t$ is determined by sampling from $p(m_t|m_{t-1}) \in \mathbb{R}^{n_m\times n_m}$, which is a 2-dimensional multinomial distribution. Note that to disallow self-transitions and ensure that mode changes to another state whenever $d_{t-1}=1$, we set all the diagonal entries in $p(m_t|m_{t-1})$ to zero.

The duration variable $d_t$ is modeled as
\begin{align}
p(d_t|m_t, d_{t-1}) &= \begin{cases} p(d_t|m_t) &\mbox{if } d_{t-1}=1 \\
\delta(d_t, d_{t-1}\hspace{-0pt}-\hspace{-0pt}1) & \mbox{if } d_{t-1} > 1 \end{cases},
\label{eq:durDist}
\end{align}
which means that as long as $d_{t-1} > 1$, we simply set $d_{t} = d_{t-1} - 1$. On the other hand, for $d_{t-1} = 1$, the new duration $d_t$ is determined by sampling from $p(d_t|m_t)$, based on the current value of mode $m_t$. In this work we assume that
\begin{align*}
p(d_t|m_t) := P(d_t = k | m_t) = \frac{\lambda_{m_t}^k e^{-\lambda_{m_t}}}{k!}
\end{align*}
is a Poisson distribution with $\lambda_{m_t} > 0$ and $k = 0,1,2,\ldots$. Observe that $d_t$ together with $m_t$ define a semi-Markov process \cite{jali13} (top two chains shown in the left plot of Figure \ref{fig:smsvar}), which has more flexibility in modeling mode durations as opposed to a simple Markov process, whose implicit mode duration is constrained to a geometric distribution \cite{ross14}.

The phase $x_t$ is distributed according to the following transition model
\begin{align}
p(x_t|x_{t-1}, m_t, d_{t-1}) &=\begin{cases} p(x_t|x_{t-1},m_t) & \mbox{if } d_{t-1}=1 \\
\delta(x_t, x_{t-1}) &\mbox{if} ~d_{t-1} > 1 \end{cases}.
\label{eq:phaseDist}
\end{align}
Here $p(x_t|x_{t-1}, m_t) \in \mathbb{R}^{n_x\times n_x \times n_m}$ is a 3-dimensional multinomial distribution of $x_t$: for each value of the mode $m_{t}$, there is a separate transition matrix defining the distribution of $x_t$, given $x_{t-1}$: $p(x_t|x_{t-1})$. It is invoked whenever the counter $d_{t-1}=1$, otherwise the phase is forced to stay in the same state. We note that in contrast to the mode distribution in \eqref{eq:modeDist}, here we allow self-transitions even when $d_{t-1}=1$, i.e., $p(x_t|x_{t-1}, m_t)$ for each $m_t$ can have non-zero diagonal. The idea behind this modeling step comes from the motivation to enable long-lasting phase persistence. In other words, although the mode can switch quickly to another state, the phase, on the other hand, has a flexibility to either remain unchanged or transition to another phase. In fact, this property of our model is precisely what enables to compress many mode states into a single phase, allowing the use of only a few VAR processes to model the data.

Finally, the sensor measurements are modeled using first-order VAR process:
\begin{align}
y_t = A_{x_t}y_{t-1} + \epsilon_{x_t},
\label{eq:VAR}
\end{align}
where $A_{x_t} \in \mathbb{R}^{n_y\times n_y}$ is the VAR transition matrix and $\epsilon_{x_t} \sim \mathcal{N}(0, \Sigma_{x_t})$ is a Gaussian noise, uncorrelated in time $t$. The probability distribution of $y_t$ then takes the form
\begin{align}
p(y_t|y_{t-1},x_t) \propto C e^{-\frac{1}{2}(y_t-A_{x_t}y_{t-1})^T\Sigma_{x_t}^{-1}(y_t-A_{x_t}y_{t-1})},
\label{eq:sensDist}
\end{align}
for some normalization constant $C$. Note that for each value of the phase variable $x_t$, there is a separate VAR process with its own transition matrix and noise characteristics, which determines the time evolution of the vector $y$. In this work we assume that $\Sigma_{x_t}$ is identity, thus the only unknown parameter in \eqref{eq:sensDist} is a transition matrix $A_{x_t}$. Each VAR process is assumed to be stable \cite{lutk07}, i.e., all the eigenvalues of $A_{x_t}$ have magnitude less than 1.

We remark on several important points about our model. (i) Note that SMS-VAR is similar but different from Markov switching autoregressive model \cite{hami89}, also known in literature as hybrid, switching state-space models or jump-linear systems \cite{ghhi00}. The main difference is that the switching dynamics is governed by a hierarchy of observed, $m_t$, and unobserved, $x_t$ semi-Markov processes, rather than a single unobserved Markov process. (ii) We could have defined our model without a phase $x_t$, where the mode $m_t$ would directly determine the active VAR process. However, whenever any of the switches change its state, there would be a transition to a different VAR. This is a bad design since the transitions would be too frequent and the number of VARs would be too large. Including phase $x_t$, which depends on mode $m_t$ and duration $d_t$, enables data compression since not every switch change would result in the phase change. This behavior is observed on the left of Figure \ref{fig:model}, where phase 5 is insensitive to a change in auto thrust at $t=50$, similarly phase 1 at $t=1200$ stays same, although some switches change. (iii) Preliminary evaluations of the flight dataset based on correlogram \cite{lutk07} revealed that sample autocorrelation functions of the time series exhibit a fast decay beyond the first lag. i.e., there are no long-range interaction between two events far away from each other. Therefore, the use of first-order dependency in our model is adequate to represent the data. Based on this, we position SMS-VAR as a short-memory model \cite{hass15} and the proposed anomaly detector in Section \ref{sec:AnomDetect} specifically targets short-term anomaly events.

\subsection{Parameter Learning}

Given data $D = \{F^1, \ldots, F^N\}$, consisting of $N$ time series $F^i = \{\bar{d}_1^i, \ldots, \bar{d}_{T_i}^i, \bar{m}_1^i,\ldots, \bar{m}_{T_i}^i, \bar{y}_1^i,\ldots,\bar{y}_{T_i}^i\}$, (bar over the variable means that it is observed), our objective here is to estimate the parameters of SMS-VAR model:
\begin{align}
\Theta = \{p(m_t|m_{t-1}), p(x_t|x_{t-1}, m_t), \lambda_{m_t}, A_{x_t}\}.
\label{eq:theta}
\end{align}

Since the data related to hidden phase $X^i = \{x_1^i,\ldots, x_{T_i}^i\}$ is unobservable, the standard approach is to use Expectation-Maximization (EM) algorithm \cite{delr77}. The idea is to find $\Theta$, which maximizes likelihood of all the observed and unobserved data $p\Big(F^{1:N}, X^{1:N}\Big|\Theta\Big)$. Assuming that we have an initial estimate of parameters $\Theta_0$, the EM algorithm consists of iterating the following two steps until convergence:


%
\begin{itemize}
\item $E$-step: $Q(\Theta, \hspace{-0pt}\Theta_{k}) \hspace{0pt}= \hspace{-0pt}\mathbb{E}_{X^{1:N}}\hspace{-0pt}\bigg[\log p\Big(F^{1:N}\hspace{-0pt}, X^{1:N}\Big|\Theta\Big)\bigg|F^{1:N}\hspace{-0pt}, \Theta_k\bigg]$
\item $M$-step: $\Theta_{k+1} = \underset{\Theta}{\arg\max}~Q(\Theta, \Theta_k)$.
\end{itemize}
Note that the $E$-step is executed for each flight independently, while in $M$-step the resulting probability information from all flights is collected to update the model parameters.

Also, observe that the execution of EM for the considered model can be challenging since the nodes in DBN are of mixed data type, complicating the inference in $E$ and optimization in $M$ steps. However, exploiting the fact that $d_t, m_t$ and $y_t$ are observable, we can compute both steps very efficiently.

\subsubsection{$E$-step}
Given the parameter specifications in Section \ref{sec:modelSpec}, the $E$-step can be efficiently computed using Junction Tree algorithm \cite{barb12}. Specifically, based on the DBN structure of the model on the left plot of Figure \ref{fig:smsvar}, we construct its junction tree (JT), shown on the right plot of Figure \ref{fig:smsvar}. JT is simply a tree-structured representation of the graph, which helps to decompose the global computations of joint probability $p\Big(F^{1:N}, X^{1:N}\Big|\Theta\Big)$ into a linked set of local computations.

Each oval node in junction tree in Figure \ref{fig:smsvar}, representing cliques in the graph of SMS-VAR, is initialized with a value of the corresponding probability distribution. For example, a node $\mathcal{D}_t$ is initialized with the value of duration distribution $p(\bar{m}_t|\bar{m}_{t-1},\bar{d}_{t-1})$ in \eqref{eq:durDist}. Similarly, $\mathcal{M}_t$, $\mathcal{X}_t$, and $\mathcal{Y}_t$ are initialized by evaluating \eqref{eq:modeDist}, \eqref{eq:phaseDist} and \eqref{eq:sensDist}  on the data.

After all the $N$ trees are initialized with the data $D = \{F^1, \ldots, F^N\}$, the operation of Junction Tree algorithm to compute $E$-step consists of propagating messages forward in time (form $t=1$ to $t = T_i$) and backward in time (form $t = T_i$ to $t=1$). For example, at iteration $k$, the result of forward propagation is the computation of likelihood of data $p(F^{1:N}|\Theta_k)$, while the backward propagation computes $p(x_t,x_{t+1}|F^{1:N},\Theta_k)$ and $p(x_t|F^{1:N},\Theta_k)$ for each $t$.

The important practical aspect of the above calculations is to prevent numerical underflow, occurring during message propagation when many small numbers multiplied together. To avoid this, we performed all operations in log-scale and at each time stamp $t$ normalized the messages to have their probability mass sum to one (using log-sum-exp function).

\subsubsection{$M$-step}

The result of $E$-step is now used to update the parameter estimates $\Theta$ in \eqref{eq:theta}. We note that since phase $x_t$ is the only unobservable part of the model, the only parameters that are re-estimated are phase transition distribution $p(x_t|x_{t-1},m_t)$ and VAR transition matrices $A_{x_t}$. The other parameters, i.e., mode and duration distributions, depend on variables which are completely observable and estimated directly from data once and never re-estimated during EM iterations.

To estimate phase transition distribution $p(x_t|x_{t-1},m_t)$, it can be shown that optimization problem \\ $\underset{\Theta}{\arg\max}~Q(\Theta, \Theta_k)$ for $p(x_t|x_{t-1}, m_t)$ amounts to estimating for each value of $m_t$ the transition matrix $p(x_t|x_{t-1})$ by multiplying matrices $p(x_t,x_{t+1}|F^1\hspace{-3pt}:\hspace{-2pt}F^N,\Theta_k)$ across those time steps $t$ at which $m_t$ matches the mode value $\bar{m}_t$ in the data.

The solution of $\underset{\Theta}{\arg\max}~Q(\Theta, \Theta_k)$  for $A_{x_t}$ can be shown to be equivalent to a solution of a least-square problem for each $x_t \in \{1,\ldots, n_x\}$. Specifically, using $p(x_t|F^{1:N},\Theta)$ from the results of $E$-step, we weight each sample vector $\bar{y}_t$ with a scalar $w_t = p(x_t|F^{1:N},\Theta)$ for one of the $x_t$: $\bar{y}_t^\prime = w_t\bar{y}_t$. Then for each weighted data sequence $\bar{y}_t^\prime, \ldots, \bar{y}_{T_i}^\prime$, $i=1\ldots,N$ we can stack the vectors as in expression \eqref{eq:VAR} in a matrix form and write the following system of equations
\begin{align*}
\begin{bmatrix}
{\bar{y}_2}^{\prime T} & \bar{y}_3^{\prime T} & \hdots & \bar{y}_{T_i}^{\prime T}
\end{bmatrix}^T
=
\begin{bmatrix}
\bar{y}_1^{\prime T} & \bar{y}_2^{\prime T} & \hdots & \bar{y}_{T_i-1}^{\prime T}
\end{bmatrix}^T
A_{x_t}^T.
\end{align*}
In compact notations above can be written as $Y_i = M_iB$, where $Y_i, M_i \in \mathbb{R}^{L_i\times n_y}$ and $B = A_{x_t}^T \in \mathbb{R}^{n_y\times n_y}$ for $L_i = T_i - 1$. Now stacking together the equations for all $N$ data sequences, we get $[Y_1^T, \ldots, Y_N^T]^T = [M_1^T, \ldots, M_N^T]^T B$, which again can be compactly written as a matrix equation $Y=MB$, where $Y, M \in \mathbb{R}^{L\times n_y}$ for $L = \sum_{i=1}^N L_i$. Now vectorizing (column-wise) each matrix in $Y=MB$, we get
\begin{align*}
\text{vec}(Y) &= (I_{n_y\times n_y} \otimes M)\text{vec}(Y)\\
\mathbf{y} &= Z \bm{\beta},
\end{align*}
where now $\mathbf{y}\in \mathbb{R}^{Ln_y}$,  $Z = (I_{n_y\times n_y} \otimes M) \in \mathbb{R}^{Ln_y\times n_y^2}$, $\bm{\beta} \in \mathbb{R}^{n_y^2}$ and $\otimes$ is the Kronecker product. Consequently, $\bm{\beta}$ (rows of $A_{x_t}$ stacked in a vector) is estimated by solving
\begin{align}
\underset{\bm{\beta}}{\arg \min}\frac{1}{L}\|\mathbf{y} - Z\bm{\beta}\|_2^2.
\label{eq:LS}
\end{align}
Note that matrix $Z$ in \eqref{eq:LS} can become very tall in cases when there are many data sequences $N$, each of large length $T_i$. The standard approaches of estimating $\bm{\beta}$, based on regular QR decomposition \cite{golo12}, become impractical. For this purpose, in practice, we use the approach of \cite{dghm12} based on Tall and Skinny QR (TSQR), which enables to perform QR of a tall matrix in a block-by-block manner.

To summarize, the execution of the above two steps can be computed in a very efficient, parallel manner. For instance, $E$ step is completely separable across flights and although $M$ step (main computation is \eqref{eq:LS}), requires synchronization, it can also be distributed with the use of parallel QR \cite{dghm12}.

\section{Anomaly Detection Framework}
\label{sec:AnomDetect}
Given  a dataset of $N$ \emph{unlabeled} flights our objective is to detect which of them deviate from the normal behavior the most. Since there is no a-priori knowledge on which flights belong to which category, we cannot build a separate model for each class. Our approach is to construct a single SMS-VAR model using \emph{all} flight data and then evaluate the built model on \emph{all} the flights to detect anomalies. Assuming that the fraction of anomalous flights in the dataset is small, we expect that the constructed model mostly represents a typical normal aircraft behavior and is not significantly influenced by the abnormal data. See Section \ref{sec:ExperimentRatio} for further discussion of this point.

Evaluation of the constructed model on the flights is a critical step of the approach since it determines the detection accuracy of the framework. One simple and straightforward choice is the computation of likelihood of each flight. For example, given estimated model parameters $\Theta$, we can compute the likelihood of the data of flight $i$ (dropping $i$ related to the numbering of time series to avoid clutter) 
\begin{align}
\label{eq:LL}
p(F) &= p(\bar{d}_{1:T}, \bar{m}_{1:T}, \bar{y}_{1:T}) = \\
& = p(\bar{d}_1, \bar{m}_1, \bar{y}_1)\prod_{t=2}^{T}p(\bar{d}_t, \bar{m}_t, \bar{y}_t |\bar{d}_{1:t-1}, \bar{m}_{1:t-1}, \bar{y}_{1:t-1} )\nonumber
\end{align}
where $\ell_t = p(\bar{d}_t, \bar{m}_t, \bar{y}_t |\bar{d}_{1:t-1}, \bar{m}_{1:t-1}, \bar{y}_{1:t-1} )$ is the likelihood of observation at $t$ given data seen so far. At each time stamp we can monitor the flight and flag down the times when probability drops to small values, signaling of anomalous activity. However, as we subsequently show in Section \ref{sec:Experiments}, this metric did not perform well as compared to the approach we propose in this work and discuss below. 

Our proposed method to monitor flight behavior is based on evaluating a discrepancy between one-step ahead predicted and the filtered phase distribution, i.e., after data observation. Specifically, assume that the current phase distribution $x_t$ over $\{1,\ldots, n_x\}$ is $p(x_t|\bar{d}_{1:t}, \bar{m}_{1:t}, \bar{y}_{1:t})$. Then we can propagate this probability one step forward using estimated model parameters to get prior estimate of phase distribution at $t+1$ (see left plot of Figure \ref{fig:smsvar} for details):
\begin{align}
\label{eq:prior}
p(x_{t+1}|\bar{d}_{1:t}, \bar{m}_{1:t}, \bar{y}_{1:t}) 
 & =\hspace{-5pt}\sum_{d_{t+1}, m_{t+1}, x_t, y_{t+1}} \hspace{-7pt} p(d_{t+1}, m_{t+1}, x_{t},x_{t+1}, y_{t+1}|\bar{d}_{1:t}, \bar{m}_{1:t}, \bar{y}_{1:t})\\
& = \hspace{-5pt}\sum_{d_{t+1}, m_{t+1}, x_t, y_{t+1}} \hspace{-7pt} p(x_t|\bar{d}_{1:t}, \bar{m}_{1:t}, \bar{y}_{1:t})~p(d_{t+1}|m_{t+1},\bar{d}_t)\times\nonumber\\
& \times p(m_{t+1}|\bar{m}_t,\bar{d}_{t})~p(x_{t+1}|x_t,m_{t+1},\bar{d}_t)~p(y_{t+1}|\bar{y}_t,x_{t+1})\nonumber.
\end{align}
At time $t+1$ we observe data $\bar{m}_{t+1}$, $\bar{d}_{t+1}$, $\bar{y}_{t+1}$, and so the posterior (filtered) distribution of the phase changes to 
\begin{align}
\label{eq:post}
& p(x_{t+1}|\bar{d}_{1:t+1}, \bar{m}_{1:t+1}, \bar{y}_{1:t+1}) = \\
& \quad \quad \quad \quad = \frac{p(\bar{d}_{t+1}, \bar{m}_{t+1}, x_{t+1}, \bar{y}_{t+1}|\bar{d}_{1:t}, \bar{m}_{1:t}, \bar{y}_{1:t})}{\sum_{x_{t+1}}p(\bar{d}_{t+1}, \bar{m}_{t+1}, x_{t+1}, \bar{y}_{t+1}|\bar{d}_{1:t}, \bar{m}_{1:t}, \bar{y}_{1:t})}\nonumber,
\end{align}
where we computed
\begin{align}
\label{eq:postDetails}
& p(\bar{d}_{t+1}, \bar{m}_{t+1}, x_{t+1}, \bar{y}_{t+1}|\bar{d}_{1:t}, \bar{m}_{1:t}, \bar{y}_{1:t}) = \\
 &\quad \quad =\sum_{x_t}  p(\bar{d}_{t+1}, \bar{m}_{t+1}, x_{t},x_{t+1}, \bar{y}_{t+1}|\bar{d}_{1:t}, \bar{m}_{1:t}, \bar{y}_{1:t})\nonumber\\
&\quad\quad = \sum_{x_t} p(x_t|\bar{d}_{1:t}, \bar{m}_{1:t}, \bar{y}_{1:t})~p(\bar{d}_{t+1}|\bar{m}_{t+1},\bar{d}_t)\times\nonumber\\
&\quad\quad \times p(\bar{m}_{t+1}|\bar{m}_t,\bar{d}_{t})~p(x_{t+1}|x_t,\bar{m}_{t+1},\bar{d}_{t})~p(\bar{y}_{t+1}|\bar{y}_t,x_{t+1})\nonumber.
\end{align}

Next, given \eqref{eq:prior} and \eqref{eq:post}, i.e., the distribution of the hidden phase before and after observations at time $t+1$, we compare these two distributions and measure their difference. For this purpose, we use Kullback--Leibler (KL) divergence, which is defined as 
\begin{align}
\label{eq:kl}
D_{t+1}\Big[ p(x_{t+1}&|F_{1:t}) \Big|\Big| p(x_{t+1}|F_{1:t+1})\Big] = \\
& = \sum_{x_{t+1}\in\{1,\ldots,n_x\}}p(x_{t+1}|F_{1:t})\log\frac{p(x_{t+1}|F_{1:t})}{p(x_{t+1}|F_{1:t+1})},\nonumber
\end{align}
where $F_{1:t}$ is a shorthand for $\{\bar{d}_{1:t}, \bar{m}_{1:t}, \bar{y}_{1:t}\}$, and similarly for $F_{1:t+1}$. Observe from \eqref{eq:prior} and \eqref{eq:postDetails} that information about all the observed variables, $\bar{d}_t$, $\bar{m}_t$ and $\bar{y}_t$ participate in the computation of the phase distribution. The prior $p(x_{t+1}|F_{1:t})$ reflects the model's belief based on data seen so far about the probability of which phase (i.e., VAR process) currently is active. After data observation, if the posterior $p(x_{t+1}|F_{1:t+1})$ shows a different phase distribution, then the distance measure \eqref{eq:kl} captures this by producing a large $D_t$. On the other hand, when both distributions are similar, it means the observed data are likely to have been generated from the model and the computed value $D_t$ is small. Moreover, observe that incremental nature of the computation of $D_t$ values implies that our approach can be used for \emph{online} anomaly detection, i.e., algorithm can monitor a flight in real time, without the need to wait for all the data to arrive.

Finally, once we compute $D_t$'s, the anomaly score $S$, characterizing the flight's abnormality, is obtained by computing a standard deviation (STD) of all $D_{t}$, i.e.,
\begin{align}
\label{eq:stdKL}
S_{\text{KL}} = \frac{1}{T}\sum_{t=1}^T(D_t-\mu_{D})^2,
\end{align}
where $\mu_{D} = \frac{1}{T}\sum_{t=1}^TD_t$. Thus, small values of $S$ indicate normal flights, while large $S$ correspond to anomalies.

To motivate our choice for using STD, note that the first-order SMS-VAR is a short-memory model (see Section \ref{sec:Model}), which usually detects short-duration anomalies. Moreover, the values $D_t$, comparing phases at neighboring times, are sensitive to short-duration anomalies, and so for a typical abnormal flight, $D_t$ usually stay small except for a few time stamps, at which anomaly events occurs and $D_t$ spike.
%
%
Therefore, STD which uses a sum of quadratic deviations, is sufficiently sensitive to outliers and can serve as an adequate summary measure. We note that detecting abnormal flights based on $D_t$ values can be viewed as a separate outlier detection problem in univariate time-series. Various sophisticated approaches can be used for this, e.g., based on support vector regression \cite{mape03}, mixture transition distribution approach of \cite{lemr96} or a median information from the neighborhood \cite{bame07}, although we observed that a simple standard deviation performed well in practice.


\section{Empirical Evaluation: Baseline Algorithms}
\label{sec:ComparedAlgos}
In this Section we present an overview of the algorithms which we used in the comparison studies in Section \ref{sec:Experiments}.The five algorithms we considered were: SMS-VAR KL, based on KL divergence (discussed in Section \ref{sec:AnomDetect}), SMS-VAR based on log-likelihood, vector autoregressive (VAR) model, semi-Markov switching (SMM) model and the multiple kernel anomaly detector (MKAD).

\textbf{SMS-VAR LL}. This approach is based on estimating SMS-VAR using all the data points and evaluating the flights based on log-likelihood of data. However, instead of using $p(F)$ from \eqref{eq:LL} directly as a measure of abnormality, which usually washes out the events of interest and makes them undetectable; similarly as in \eqref{eq:stdKL} we used standard deviation, i.e., $S_{\text{LL}} = \frac{1}{T}\sum_{t=2}^T(\log \ell_t - \mu_{\ell})^2$, where $\ell_t = p(\bar{d}_t, \bar{m}_t, \bar{y}_t |\bar{d}_{1:t-1}, \bar{m}_{1:t-1}, \bar{y}_{1:t-1} )$ and $\mu_{\ell}$ is the mean of $\ell_t$'s.

\textbf{VAR}. In this approach we modeled only the continuous part of the data using a single first-order VAR process $y_t = Ay_{t-1} + \epsilon_t$. The anomaly detection is based on computing $S_{\text{VAR}} = \frac{1}{T}\sum_{t=2}^T(r_t - \mu_{r})^2$, where $r_t = \|y_t - \hat{A}y_{t-1}\|_2$ is a one-step-ahead prediction error, $\hat{A}$ is the estimated VAR matrix using all the data and $\mu_{r}$ is the mean. 

\begin{figure}
\centering
\includegraphics[width=0.4\textwidth]{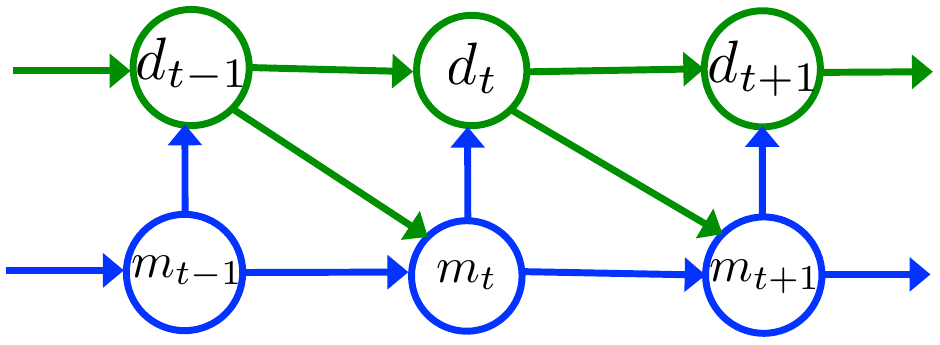}
\caption{Semi-Markov model to represent discrete flight data related to pilot switches.}
\label{fig:smm}
\end{figure}

\textbf{SMM}. An approach based on modeling only the discrete part of data is based on semi-Markov model shown in Figure \ref{fig:smm}. Here, similarly as in SMS-VAR, we modeled $m_t$ and $d_t$ using \eqref{eq:modeDist} and \eqref{eq:durDist}, respectively. The anomaly score is then computed as $S_{\text{SMM}} = \frac{1}{T}\sum_{t=2}^T(\ell_t - \mu_{\ell})^2$, where $\ell = p(\bar{d}_t, \bar{m}_t|\bar{d}_{1:t-1}, \bar{m}_{1:t-1})$.

\begin{figure*}[!t]
\centering
\includegraphics[width=\textwidth]{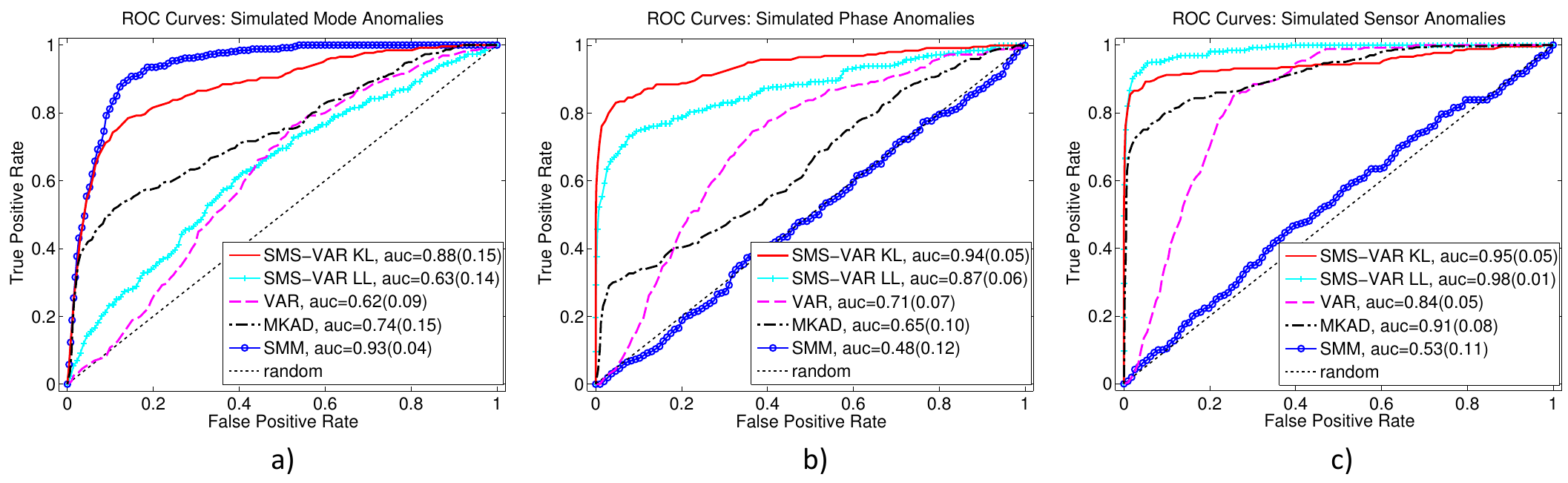}
\caption{Performance of the five algorithms at detecting three simulated types of anomaly events: a) Mode anomaly, b) Phase anomaly, c) Sensor anomaly. There were 100 normal and 10 anomalous flights in each considered scenario. The evaluation was done using the area under ROC curve (AUC). The AUC values are shown after averaging the results over 30 runs, the values in parenthesis show one standard deviation. }
\label{fig:sim}
\end{figure*}

\textbf{MKAD}. This algorithm was designed to detect anomalies in the heterogeneous multivariate time series, where both discrete and continuous features are present. Let $F^i$ and $F^j$ denote the multivariate time series of two flights. The algorithm constructs a kernel of the form $K\Big(F^i, F^j\Big) = \alpha K_d\Big(F^i, F^j\Big) + (1-\alpha)K_c\Big(F^i, F^j\Big)$, where $K_d$ is a kernel over discrete sequences and $K_c$ is a kernel over continuous time series and $\alpha \in [0,1]$ is a weight, usually set to $\alpha = 0.5$. For discrete sequences, the normalized longest common subsequence (LCS) is used, i.e., $K_d\Big(F^i, F^j\Big) = \frac{|LCS(F^i, F^j)|}{\sqrt{T_iT_j}}$, where $|LCS(F^i, F^j)|$ denotes the length of LCS. For continuous sequences, the kernel $K_c\Big(F^i, F^j\Big)$ is inversely proportional to the distance between symbolic aggregate approximation (SAX) representation \cite{pkll02} of continuous sequences in $F^i$ and $F^j$. The constructed kernel $K \in \mathbb{R}^{N\times N}$, $N$ is the number of flights, is then used in one-class support vector machine (SVM) to construct a hyperplane to separate anomalous and normal flights. Note that the main idea behind SAX technique is to represent a continuous sequence as a discrete one. This is achieved by dividing a sequence into equally spaced segments, computing the average of each segment and then discretizing the result into a set of alphabets of predefined size. By regulating the length of the segments, MKAD can be tuned to detect long- or short-term dependencies in the data.

\section{Experimental Results}
\label{sec:Experiments}
In this Section we present extensive evaluations of the SMS-VAR model on both synthetic as well as real flight data. We compare the performance of the proposed approach with the four alternatives described in Section \ref{sec:ComparedAlgos}.

\subsection{Synthetic Data}
In the first simulation scenario we study detection accuracy of all the algorithms when presented with different types of anomalies, while in the second study we examine how the proportion of anomalous to normal flights affects the detection accuracy of SMS-VAR algorithm.

\subsubsection{Detecting Different Types of Anomalies}

\begin{figure}[!t]
\centering
\includegraphics[width=.5\textwidth]{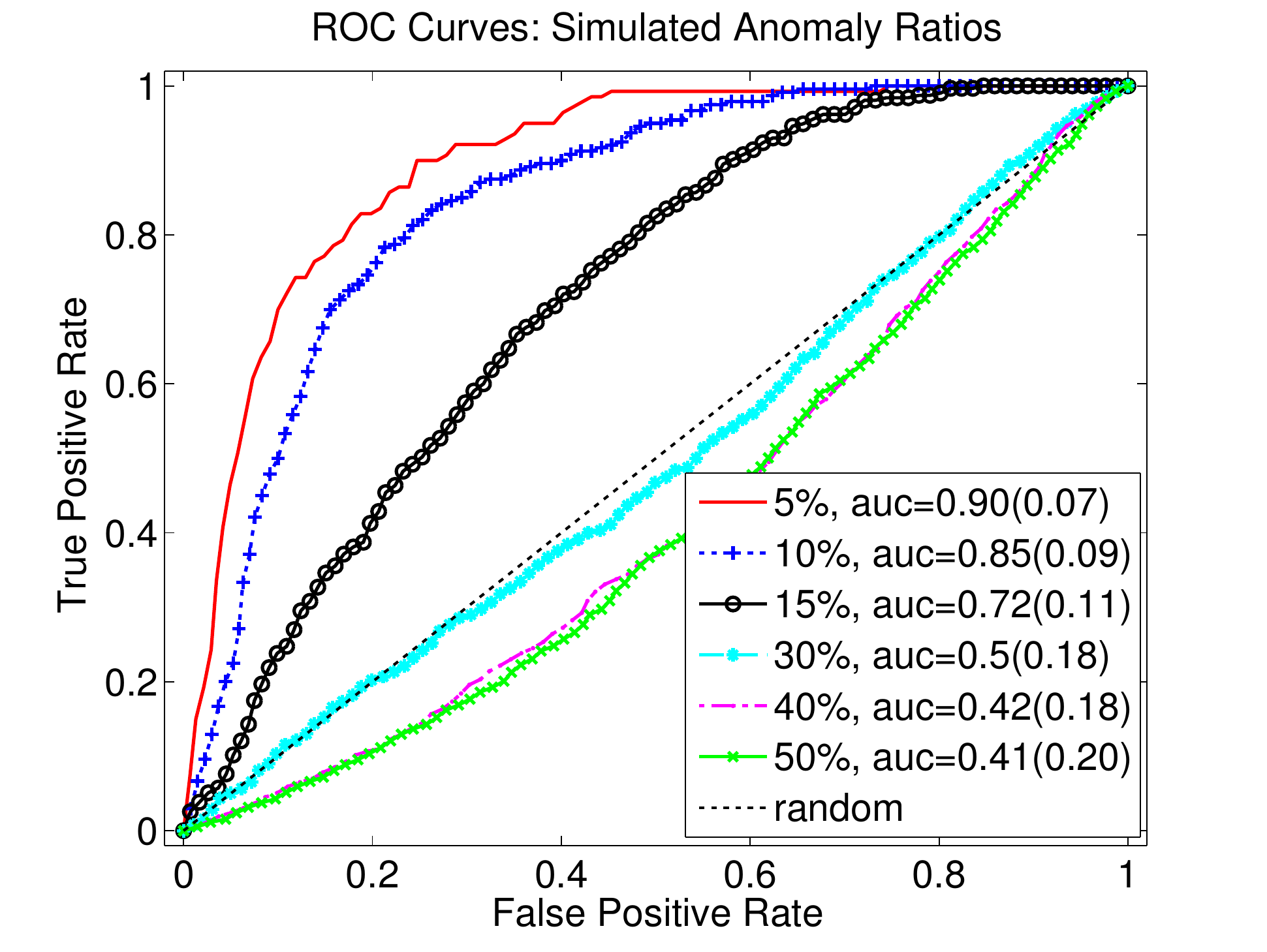}
\caption{Anomaly detection accuracy of SMS-VAR KL when applied to datasets with different proportion of anomalous data ($5\%-50\%$ of normal data is replaced with anomalous). Total number of flights in all scenarios is $100$.}
\label{fig:simRatios}
\end{figure}

We generated three synthetic datasets consisting of 100 normal flights and 10 anomalous, each of length 200 time stamps. The number of phases and sensor measurements in the generation model was set, respectively, to $n_x = 3$ and $n_y = 4$. The number of binary switches was set to 5, which corresponds to $n_m = 2^5 = 32$. In each dataset, the 10 abnormal flights represented a different type of anomaly. In the first dataset, each anomalous flight contained several mode anomalies, i.e., unusual flips of switches, while the continuous sensor measurements behaved normally. In the second dataset, we simulated unusual change of unobserved phases, i.e., at several time stamps one VAR process switched to another VAR process. Note that in this scenario all the measurements related to a particular VAR process were normal except that the change in underlying dynamics was abnormal. Finally, in the third dataset we simulated errors in sensor measurements (continuous data), while the mode transitions behaved normally.

The results are shown in Figure \ref{fig:sim}. It can be seen from Figure \ref{fig:sim}-a that the discrete anomaly detector SMM had the highest accuracy for detecting mode anomalies, followed by SMS-VAR and others. Since the dataset contained only discrete anomalies, it is expected that SMM performed the best. It was also expected that SMM would do poorly on the other two datasets since it could not use the information from sensor measurements. Similarly, the VAR model-based anomaly detector did not do well on the first dataset in Figure \ref{fig:sim}-a but improved its accuracy in Figures \ref{fig:sim}-b, c. MKAD algorithm, using both discrete and continuous data, achieved medium level accuracy in detecting discrete anomalies and did well on the dataset with errors in sensor measurements (we used a SAX window of size $2$, since anomalies are of short duration). The algorithm, on the other hand, was insensitive to unusual phase changes in the data.

In contrast, the SMS-VAR-based algorithm, which fuses information from both sources of measurements is able to detect the anomalous flights with high accuracy across all the datasets. Moreover, the anomaly detection approach based on KL measure performed better than the log-likelihood measure (LL) on first and second datasets and did slightly worse on the third one, confirming that it is a viable method to detect abnormal activities. 

\begin{figure*}[!t]
\centering
\includegraphics[width=\textwidth]{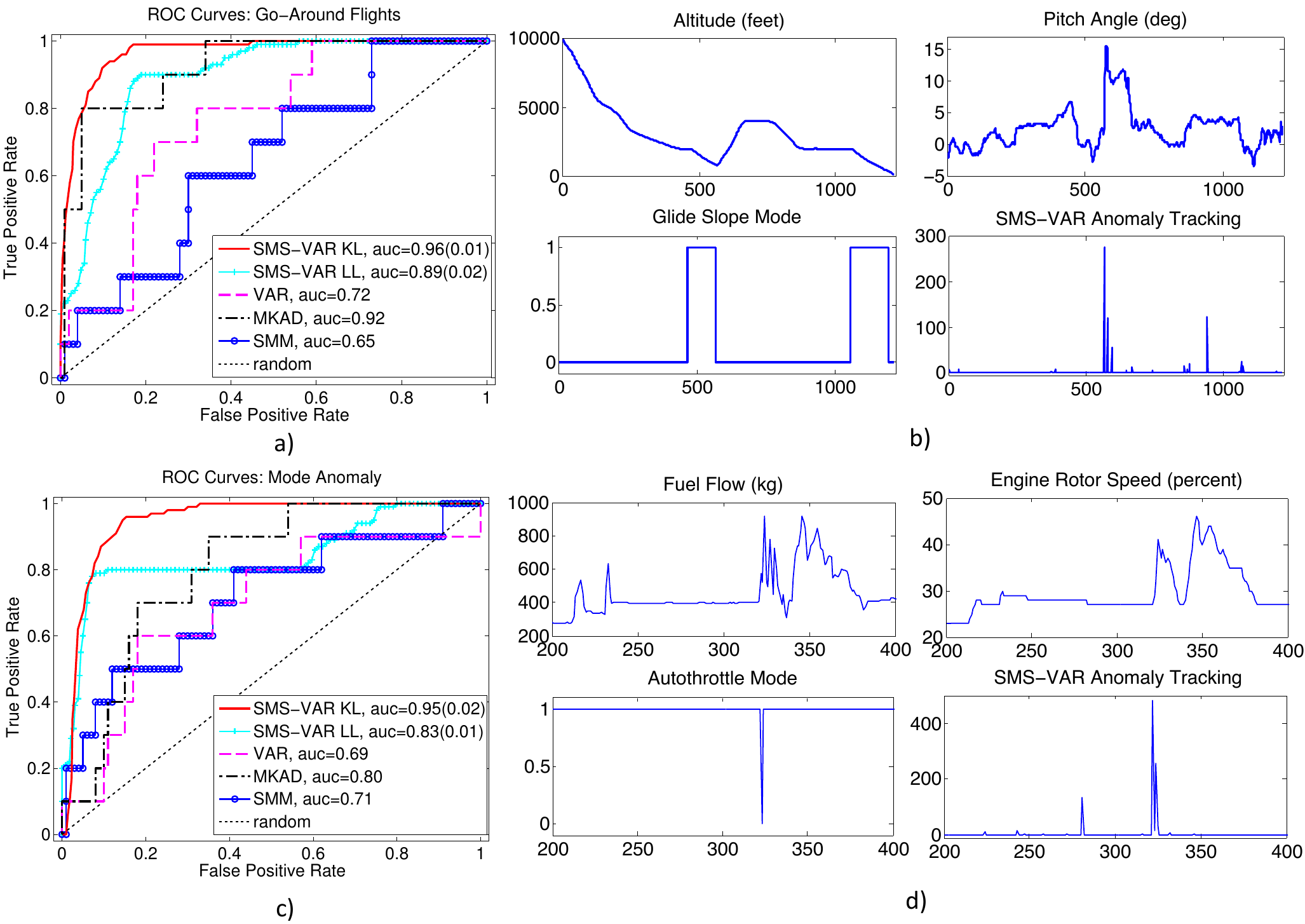}
\caption{Performance of the five approaches at detecting go-around (top) and mode anomaly (bottom) flights. There were 100 regular and 10 anomalous flights in each dataset. Figures a) and c) show ROC curves for each method. AUCs shown after averaging over 30 runs for SMS-VAR, for other methods no averaging is done since they are deterministic. Figure b) shows an example of SMS-VAR KL detecting a typical go-around flight with a history of some of the parameters. Figure d) shows detection of a typical mode anomaly flight.}
\label{fig:realGaMode}
\end{figure*}


\subsubsection{Effect of Anomaly Proportion on Accuracy}
\label{sec:ExperimentRatio}

In this study we investigate the validity of the assumption made in Section \ref{sec:AnomDetect}, where we mentioned that for our anomaly detection approach to be accurate, the fraction of anomalous flights should not be large. For this purpose we generated a set of datasets of same size but with different proportion ($5\%-50\%$) of anomalous sequences. The simulated anomaly was related to unusual switch changes; the parameter dimensions in the generation model remained the same as above. Results are shown in Figure \ref{fig:simRatios}. It can be seen that when the number of irregular data sequences is small ($5\%-15\%$), the algorithm's detection accuracy is high (AUC value is $0.9-0.7$). On the other hand, as we replace more and more normal flights with anomalous, the accuracy drops. This can be explained by noting that SMS-VAR is built using all the data, so when the fraction of anomalous flights is large, the constructed model represents now a typical \emph{anomalous} behavior. For example, in Figure \ref{fig:simRatios} we see that when $50\%$ or more flights are anomalous, the algorithm prediction flips and it starts classifying normal flights as anomalous.

\subsection{Real Flight Data}

In this section we present the evaluation results of the considered approaches on the FOQA flight dataset from a partner airline company (a similar, publicly available flight dataset can be found at \cite{nasadata}). The data contains over a million flights, each having a record of about 300 parameters, including sensor readings, control inputs and weather information, sampled at $1$ Hz. Out of 300 parameters, we selected $31$ continuous ones related to aircraft's sensor measurements and $18$ discrete binary parameters related to pilot switches. Therefore, the model dimensions are $n_y=18$ and $n_m=2^{18}$. Note that although 18 binary features correspond potentially to $2^{18} = 262144$ unique combinations, however the number of such values in real data is much smaller, since only few combinations of binary flight switches are possible. For example, in the dataset consisting of 20000 flights, used in Section \ref{sec:experRealUnlabeled}, only 673 unique modes exist ($0.3\%$ of $262144$). The number of hidden phases was set to $n_x=5$, after some preliminary experiments in which we tested several $n_x$ in range $[3, 20]$ and selected $n_x$ balancing a good prediction accuracy with low computational complexity of algorithm.  We have selected flights with landings at the same destination airport with aircrafts of the same fleet and type, so that we eliminate potential differences related to aircraft dynamics or landing patterns. Data analysis focused on a portion of the flight below 10000 feet until touchdown (duration 600-1500 time stamps), usually having the highest rates of accidents \cite{boeing}.

The evaluation of the algorithms is first done by using two small dataset of manually labeled flights (Section \ref{sec:realLabeled}), while in the second study we use a large dataset in a more realistic scenario with no information about flights' labels (Section \ref{sec:experRealUnlabeled}). In the second case the shown results are only qualitative since no ground truth is available but the discoveries were validated by the domain experts.

\begin{figure*}[!t]
\centering
\includegraphics[width=\textwidth]{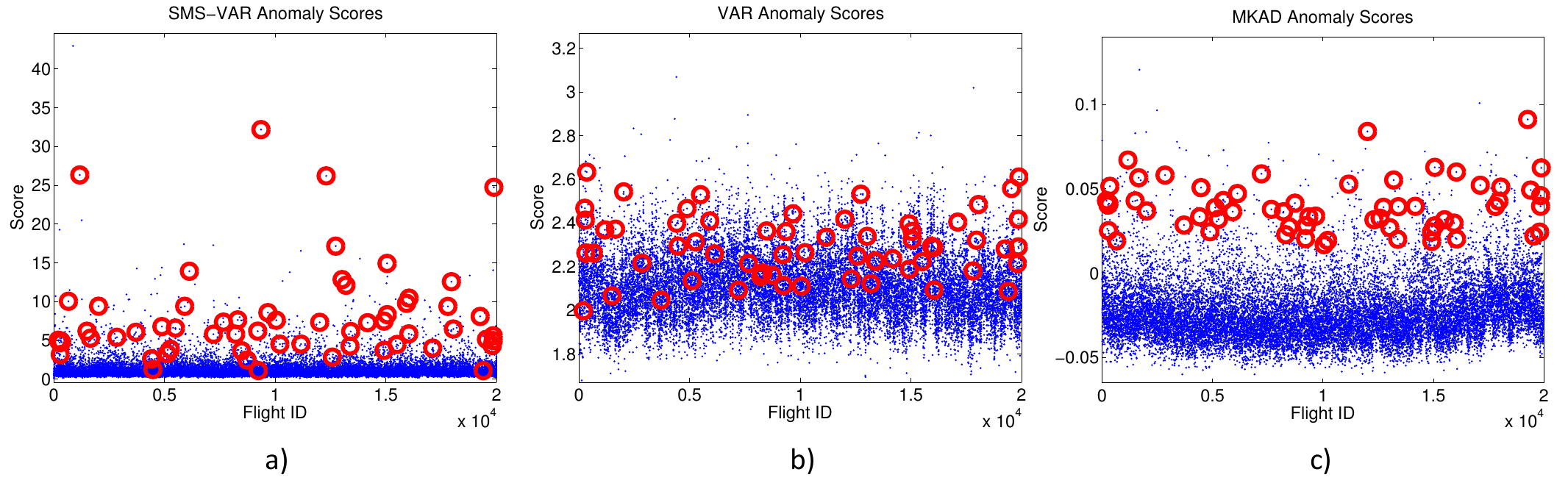}
\caption{Anomaly scores of SMS-VAR KL, VAR and MKAD on unlabeled dataset consisting of 20000 flights. Red circles in all plots denote all the 61 go-around flights present in the 20000 flights.}
\label{fig:real20000}
\end{figure*}

\begin{table*}[!t]
\centering
\begin{tabular}{|c|c|c|} 
\hline
SMS-VAR KL & VAR & MKAD \\
\hline
\hline
go-around (19) & go-around (3)  & go-around (17) \\
high speed in approach (5) &  high speed in approach (5)  & high speed in approach (2) \\
high rate of descent (4)  & fast approach (2)  & high pitch at touch down (1)    \\
bounced landing (2) & high rate of descent (4)  & low speed at touch down (1) \\
delayed braking at landing(2) & bank cycling in approach(2) & low path in approach (1)  \\
late retraction of landing gear (4) & high pitch at touch down (1) & flaps retracted in approach (1) \\
deviation from glide-slope (2) &  & unusual flight switch changes (15) \\
unusual flight switch changes  (11) &  & \\
\hline
\end{tabular}
\caption{Anomalies discovered in the top $100$ anomalous flights, ranked by each anomaly detection method in the set of $20000$ unlabeled flights. The distribution of anomaly scores for each method is shown in the Figure \ref{fig:real20000}.}
\label{tb:table1}
\end{table*}

\subsubsection{Labeled Flights}
\label{sec:realLabeled}

\textbf{Go-Around}. For this study we manually selected flights, which abort their normal landing, fly back up to a certain altitude and try to repeat the landing again (see Figure \ref{fig:model} for an example). These flights are considered operationally significant anomalies since they could be executed in response to an emergency or unsafe conditions in the air or on the runway. The dataset included 10 go-around and 100 normal flights. The results are shown in Figure \ref{fig:realGaMode}. In particular, from Figure \ref{fig:realGaMode}-a we see that SMS-VAR and MKAD performed similar, with SMS-VAR based on KL measure achieving the highest accuracy of detection. Both SMM and VAR-based approaches, which used only part of data, performed worse, missing many go-around flights. 

In Figure \ref{fig:realGaMode}-b we also show a typical go-around flight and the corresponding history of $D_t$ (see definition in \eqref{eq:kl}) values across the flight. Note that the go-around happens shortly after $t=500$, when the altitude increases to about 5000 feet, coupled with unusual behavior in other parameters. For example, a glide slope mode was switched off, which typically is used to safely descend aircraft to a runway. Also, there was a sharp rise of pitch angle, corresponding to airplane's nose lifting up during ascent. The SMS-VAR KL model correctly detected these unusual changes with multiple spikes around the time the go-around was initiated. 

\textbf{Mode Anomaly}. Additionally, we tested the performance of the algorithms on the real flight anomaly related to unusual auto-throttle switchings, whose example is shown in Figure \ref{fig:realGaMode}-d. In particular, around $t=400$ during aircraft's descent, one of the switches related to throttle control is switched off briefly ($2-5$ seconds). This action led to a fast spool up of engines (from $25\%$ to $40\%$ within few seconds). This abnormal behavior usually causes a quick increase of longitudinal acceleration leading to an abrupt forward motion of the aircraft. Similarly as before, we created a dataset consisting of 10 anomalous and 100 normal flights and the results are shown in Figure \ref{fig:realGaMode}-c. 

Interestingly, although the anomaly type was discrete, the simple SMM approach, which looks only at discrete part of data, did not perform well.  Similarly, using only continuous features, the VAR algorithm also did poorly. When the information from both sources is combined, as was done in SMS-VAR  KL or MKAD, the detection accuracy increased. Still, it can be seen that SMS-VAR KL performed better than other methods by a margin, justifying our proposed approach to track anomalies using the phase information.

\subsubsection{Unlabeled Flights}
\label{sec:experRealUnlabeled}
Finally, we compared the algorithms' performance on a dataset containing 20000 unlabeled flights. We tested SMS-VAR KL and compared its performance with MKAD and VAR. Figure \ref{fig:real20000} shows the anomaly scores for the three approaches. For each method, we examined the top 100 flights with the highest anomaly scores to determine the flights with operationally significant events. In Table \ref{tb:table1} we present a summary of the discovered anomalies, examined and validated by the experts. Note that since there is no ground truth available, the presented results are only qualitative.

As can be seen, among the top 100 flights, we found that the most common type of anomaly were the go-around flights, shown as red circles in Figure \ref{fig:real20000}, as well as anomalies related to unusual pilot switches. The number of go-arounds detected by SMS-VAR and MKAD was similar, 19 and 17, respectively. The VAR-based approach only identified 3 such flights in its 100 top anomalous flight list. SMS-VAR and MKAD algorithms have additionally identified many flights with unusual changes in pilot switches, although these flights did not overlap. In particular, the anomalies identified by SMS-VAR are characterized by quick changes in the flight parameters (few seconds), e.g., the switchings in auto throttle system as in Figure \ref{fig:realGaMode}-d or the flights when the localizer switch was turned off during approach resulting in large deviation from glide slope and these are difficult to detect using MKAD. The discrete anomalies identified by MKAD are of longer duration. For example, it found several flights when the flight director was switched off for over 2 minutes during the approach. It is an unusual behavior since, typically, flight director is used throughout the approach to assist the pilot with vertical and horizontal cues. Therefore, we can conclude that although achieving better performance in detecting certain types of anomalies, the proposed framework can be positioned as complementary to the existing state-of-the-art approaches, enabling the discoveries of more diverse spectrum of operationally significant events.


\section{Conclusions}
\label{sec:Conc}
In this work we presented a novel framework for identifying operationally significant anomalies in the aviation systems based on representing each flight as SMS-VAR model. The anomaly detection is performed using a proposed approach comparing the phase distribution predicted by the model and the one suggested by the data. Extensive experimental tests have shown that the proposed approach has high detection accuracy of various aviation safety events. Moreover, the comparison to several algorithms showed that the proposed approach outperforms many baseline methods, it also achieves similar or better detection accuracy than MKAD, identifying itself as complimentary to current state-of-the-art methods. Going forward, we plan to extend our anomaly detection approach to include the ability to model long-duration anomaly events.

\vspace*{3mm}
{\bf Acknowledgements:}
The research was supported by NASA Cooperative Agreement NNX12AQ39A, NSF grants IIS-1447566, IIS-1422557, CCF-1451986, CNS-1314560, IIS-0953274 and IIS-1029711.

\bibliographystyle{abbrv}
\bibliography{main}

\end{document}